\pdfoutput=1
\documentclass[11pt]{article}

\usepackage[]{acl}
\usepackage{times}
\usepackage{latexsym}
\usepackage{comment}
\usepackage[T1]{fontenc}

\usepackage[utf8]{inputenc}

\usepackage{microtype}

\usepackage{inconsolata}

\usepackage{graphicx}
\usepackage{latexsym}
\usepackage{epsfig}
\usepackage{graphicx}
\usepackage[linesnumbered,ruled,vlined]{algorithm2e}
\usepackage{xcolor}

\usepackage{multirow}
\usepackage{etoolbox}
\usepackage{balance}
\usepackage{balance}
\usepackage{pifont}
\usepackage{makecell}
\usepackage{hhline}
\usepackage{booktabs}
\usepackage{makecell}
\usepackage{graphicx}
\usepackage{mathtools}
\usepackage{enumitem}
\usepackage{float}
\usepackage{colortbl}
\usepackage[most]{tcolorbox}
\usepackage{pifont}
\usepackage{subcaption}
\usepackage{hyperref}
\usepackage[table]{xcolor}
\usepackage{mathtools,amssymb}
\usepackage{multirow}
\usepackage{booktabs}
\usepackage{amsmath}
\usepackage{comment}
\usepackage{amsfonts}
\usepackage[most]{tcolorbox}
\usepackage{amsmath}

\definecolor{gold}{rgb}{1.0, 0.84, 0.0}

\newcommand{\ours}{R2U\xspace}

\newcommand{\cmark}{\ding{51}}%
\newcommand{\xmark}{\ding{55}}%
\newcommand{\tmark}{\ding{115}} 

\makeatletter
\newcommand{\thickhline}{
    \noalign {\ifnum 0=`}\fi \hrule height 1pt
    \futurelet \reserved@a \@xhline
}
\newcolumntype{"}{@{\hskip\tabcolsep\vrule width 1pt\hskip\tabcolsep}}
\makeatother

\makeatletter
\newcommand\footnoteref[1]{\protected@xdef\@thefnmark{\ref{#1}}\@footnotemark}
\makeatother
\newcolumntype{P}[1]{>{\centering\arraybackslash}p{#1}}

\title{Relevance to Utility: Process-Supervised Rewrite for RAG}

\newcommand\correspondingauthor{\thanks{Corresponding author.}}
\author{Jaeyoung Kim\textsuperscript{\rm 1}\thanks{~~Equal contribution. Author order is decided by coin toss.}, Jongho Kim\textsuperscript{\rm 1}\footnotemark[\value{footnote}]
Seung-won Hwang\correspondingauthor\textsuperscript{1}, 
Seoho Song\textsuperscript{2}, 
Young-In Song\textsuperscript{2} \\
IPAI, Seoul National University\textsuperscript{1}, Naver Corp\textsuperscript{2} \\ 
\texttt{\{jae.young, jongh97, seungwonh\}@snu.ac.kr} \\
\texttt{\{song.seoho, song.youngin\}@navercorp.com} \\
}

\begin{document}
\maketitle

\begin{abstract}
Retrieval-augmented generation systems often suffer from a gap between optimizing retrieval relevance and generative utility. With such a gap, retrieved documents may be topically relevant but still lack the content needed for effective reasoning during generation.
While existing bridge modules attempt to rewrite the retrieved text for better generation, we show how they fail by not capturing ``document utility''.
In this work, we propose \ours, with a key distinction of approximating true utility through joint observation of rewriting and answering in the reasoning process.
To distill, \ours scale such supervision to enhance reliability in distillation. 
We further construct utility-improvement supervision by measuring the generator’s gain of the answer under the rewritten context, yielding signals for fine-tuning and preference optimization.
We evaluate our method across multiple open-domain question-answering benchmarks. The empirical results demonstrate consistent improvements over strong bridging baselines
~\footnote{Source code: \url{https://anonymous.4open.science/r/R2U-4307/README.md}}.
\end{abstract}

\section{Introduction}
Large language models (LLMs)~\cite{achiam2023gpt,bai2023qwen} have demonstrated strong generative capabilities across a wide range of tasks, yet they continue to struggle with queries that require knowledge beyond their parametric memory.
Retrieval-augmented generation (RAG) addresses this limitation by retrieving external documents and conditioning generation on them.
However, a persistent gap remains between retrieval relevance and generative utility: documents that are topically relevant do not necessarily support effective reasoning during answer generation~\cite{ke2024bridging}.

To bridge this gap, recent work has introduced intermediate modules between retrieval (R) and generation (G) that rewrite retrieved documents to better support downstream generation.
Such \textbf{bridge} modules include document reranking~\cite{pradeep2023rankzephyr, yang2025rank, bge_embedding} constrained to reorder documents without modifying their content, and query-focused summarization (QFS)~\cite{xu2023recomp, li2024refiner,yoon2024compact,hwang2024exit, edge2024local, chen2025pathrag} allowed to prune or enrich content selectively.
We abstract all these as \textbf{rewriting} of transformations from retrieved documents $D$ into $D'$, considering $q$ as well.

\begin{figure}[t]
    \centering
    \includegraphics[width=1\linewidth]{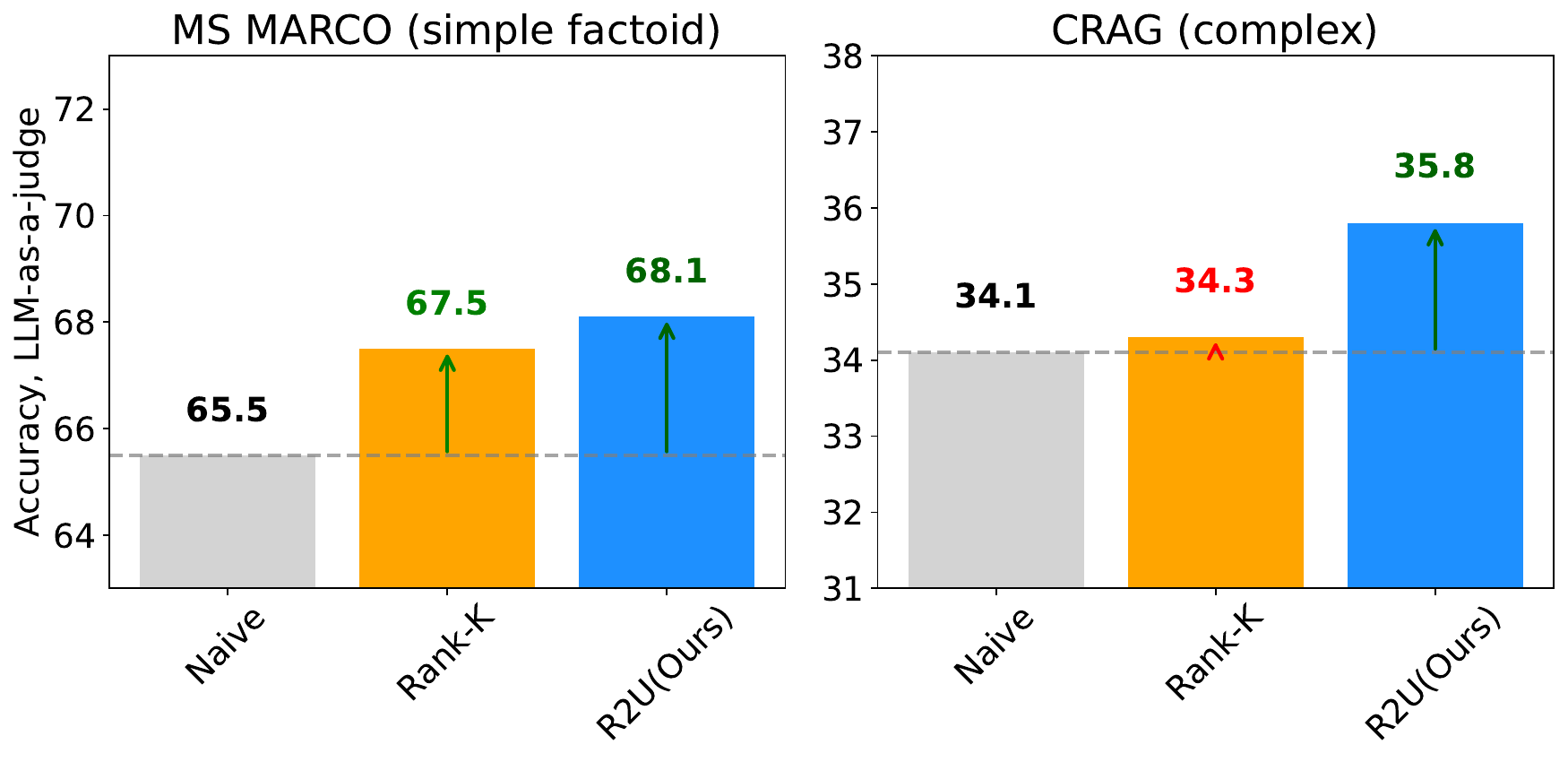}
    \caption{Accuracy gap between MS MARCO and CRAG across different bridging models. All models take the top-10 retrieved documents as input, rewrite them, and pass the results to the generator.}
    \vspace{-5mm}
    \label{fig:msmarco_vs_crag}
\end{figure}

Despite their reported success, we argue that existing rewriting approaches fail to reliably capture \textbf{document utility}, which we define as ``the extent to which a document improves a generator’s ability to produce the correct answer''.
As a result, rewriting models are particularly detrimental when the generation task requires reasoning.
For example, Figure~\ref{fig:msmarco_vs_crag}
shows 
Rank-K~\cite{yang2025rank}\footnote{Other baseline results are in Table~\ref{tab:main_results_msmarco_crag_8b}, Rank-K shows the highest average performance among the baselines.}
improves performance
on MS MARCO~\cite{DBLP:journals/corr/NguyenRSGTMD16}, where the answer is simply extracted from the documents, while it suffers on CRAG~\cite{yang2024crag}, where generation requires beyond extraction.

We attribute this failure to two fundamental limitations.
First, their rewriting process, independent of the reasoning process to derive the answer, favors shortcut documents that appear to support simple answer extraction.
For complex queries, these shortcuts mislead generators by presenting superficial matches while discarding the evidence required for reasoning steps.
Second, the binary answer for supervision is unreliable
for reasoning-heavy tasks, as a rewrite might provide crucial partial evidence yet still result in an incorrect answer due to downstream generation errors. Binary supervision discards these high-utility yet failed attempts.

To address the limitations, we propose \textbf{R}elevance \textbf{to} \textbf{U}tility (\ours), a training framework that addresses both limitations:
Rather than treating rewriting as a disjoint preprocessing step, we view it as part of the reasoning trajectory that leads to an answer.
This motivates a joint observation of rewriting and generated answer, providing a more principled utility proxy for rewriting.

Building on this formulation, \ours distills utility-aligned supervision.
First, we collect process supervision via joint rewrite–answer trajectories.
Instead of naive generation, we scale the process to isolate individual rewrites, disentangling the utility of each document from the trajectories.
Second, we improve the quality of supervision by explicitly measuring how rewritten documents change a downstream generator’s likelihood of producing the correct answer.
These signals are then used to train a lightweight rewriter via standard supervised fine-tuning and preference optimization.

We evaluate our approach on multiple complex question-answering tasks covering diverse query types and reasoning requirements.
The results show that \ours improves the average F1 score by 6.8\%pt over the naive RAG baseline and outperforms the best existing baselines by 5.6\%pt, consistently enhancing performance across tasks.

\section{On Bridging Objectives}
In this section, we formalize the role of bridge modules in RAG system and identify two fundamental limitations in existing rewriting-based approaches.
These limitations motivate the design choices of our training framework, which we address explicitly in Section \ref{subsec:proposed_approach}.

\subsection{Motivation: Existing Bridging Models}
\label{subsec:task_formulation}
In a standard RAG system, a generator $\phi$ produces an answer $a$ given a query $q$ and a set of retrieved documents $d$:
\begin{equation}
    a \sim P_{\phi}(a \mid q, d).
\end{equation}

A line of recent work introduces a bridge module that rewrites or transforms the retrieved documents to improve their utility for generation.
Such approaches include document reordering~\cite{pradeep2023rankzephyr, bge_embedding}, content filtering~\cite{yoon2024compact, li2024refiner}, and structured reconstruction~\cite{chen2025pathrag}.
These methods can be summarized as:
\begin{equation}
\label{eq:prev_bridge}
    a \sim P_{\phi}(a \mid q, d'), ~\text{where}~d' \sim P_{\theta}(d' \mid q, d),
\end{equation}
where $\theta$ denotes a rewriting module and $d'$ is the rewritten document set.

Training such bridge modules typically relies on result-level signals.
In particular, rewritten documents are sampled and evaluated based on whether the generator produces the correct answer:
\begin{equation}
\begin{split}
    \max_{\theta}~ &\mathbb{E}_{d' \sim P_{\theta}(d' \mid q,d)} \left[ \mathbf{1}(a = a^*) \right], \\
    &\text{where}~ a \sim P_{\phi}(a \mid q, d').
\end{split}
\label{eq:result_sup}
\end{equation}
We refer to this paradigm as \textbf{answer-level supervision}, since documents are rewritten relevant to the query and judged solely by the final correctness of the generated answer.
While effective in practice, this formulation suffers from two fundamental limitations that prevent it from reliably capturing \textbf{document utility improvement}, which we define as below.
\begin{tcolorbox}[title=Document utility improvement]
The extent to which a rewritten document shifts the generator’s distribution toward the correct answer
\end{tcolorbox}

The first limitation is the disjoint approximation of the document utility.
The objective of document rewriting is not merely to paraphrase the original content, but to generate documents that explicitly facilitate the reasoning process for answer generation.
However, most existing approaches approximate this objective in a decoupled manner. Rewritten documents are first generated from $p(d' \mid q, d)$ and only afterward filtered using the answer.
As a result, it is challenging to determine whether a rewritten document contributes to generating the answer or merely co-occurs with successful generation due to generator shortcuts or parametric knowledge.

The second limitation is the binary answer as a preference signal.
Such binary supervision fails to differentiate rewrite quality, obscuring the degree to which a document enhances the generator’s reasoning.
It treats all correct answers equally, ignoring cases where a rewrite actually lowers the generator's confidence yet results in a correct prediction due to model robustness.
Conversely, a rewritten document that substantially increases the probability of the correct answer may be discarded simply because it failed to cross the decision threshold for the final generation. Consequently, this binary criterion leads to the inclusion of noisy data or the exclusion of high-utility supervision signals.

\begin{figure*}[ht]
    \centering
    \includegraphics[width=0.95\linewidth]{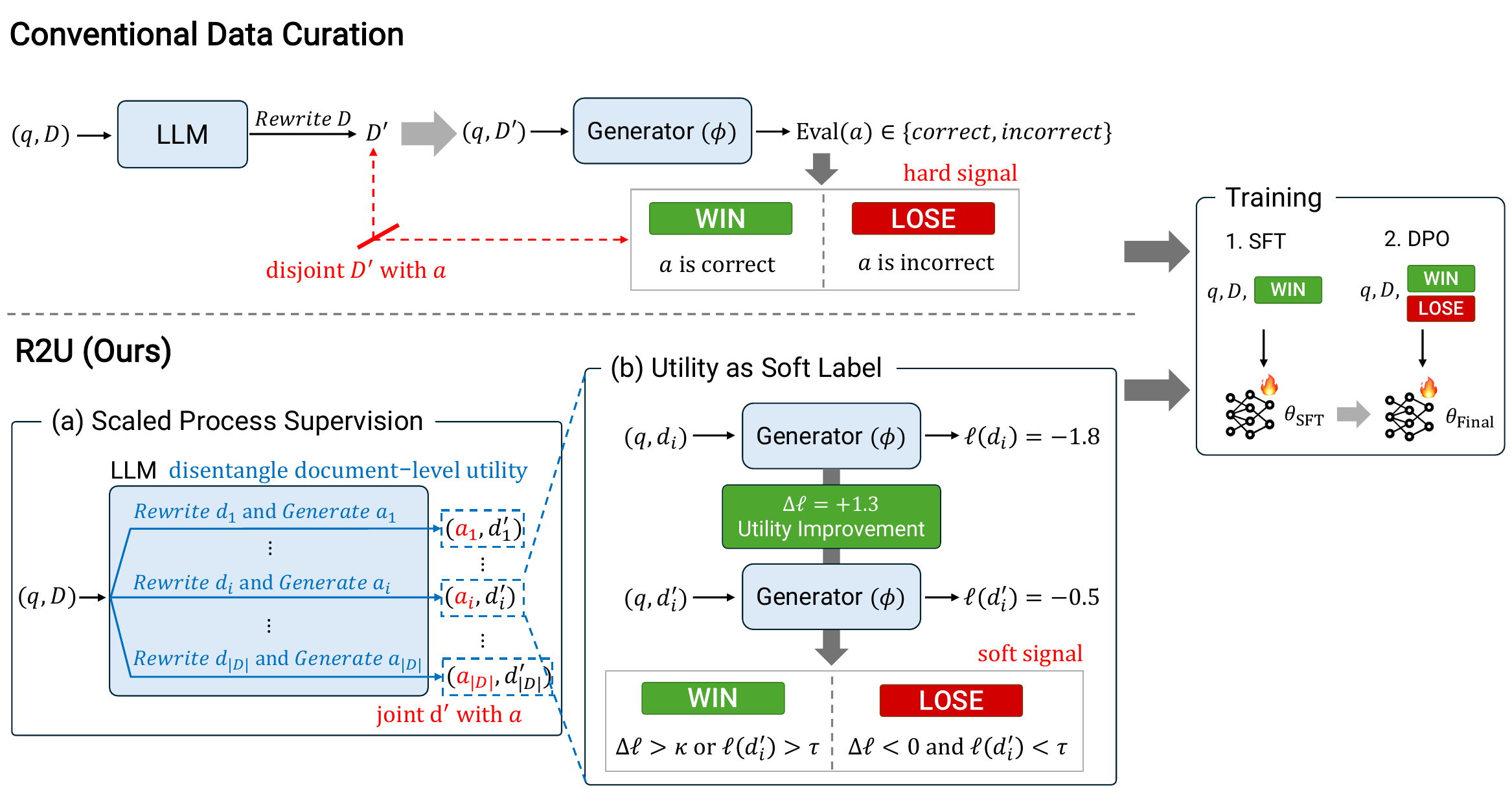}
    \caption{Overview of \ours. (a) Scaled Process Supervision: LLM rewrites each document while conditioning on the remaining documents, yielding $(a,d')$ pairs suitable for distillation. (b) Utility as Soft Label: Rather than relying on hard binary signals, we evaluate $d'_i$ by measuring their soft utility improvement, and then divide $D$ into win and lose for training rewriter via standard SFT and DPO.}

    \label{fig:ours_overview}
    \vspace{-5mm}
\end{figure*}

\section{Proposed Approach: \ours}
\label{subsec:proposed_approach}

To address the two limitations identified in Section~\ref{subsec:task_formulation}, we first clarify the objective that a document rewriting framework should satisfy.
Specifically, rewritten documents should be (i) jointly generated with the answer, and (ii) selected based on soft improvement of the downstream generation.
In the following sections, we formalize this objective from a joint perspective (\S\ref{subsec:formulation_of_ours}), and then describe how we operationalize it via scaled process supervision (\S\ref{subsec:inference_scaling}) and utility improvement (\S\ref{subsec:utility_supervision}).

\subsection{Formulation of R2U: Joint Perspective}
\label{subsec:formulation_of_ours}
Our key distinction is to treat document rewriting as part of the reasoning process that produces the answer, rather than as an independent preprocessing step.
Accordingly, we model generation as a joint distribution over rewritten documents $d'$ and the answer $a$: $P_{\phi}(a, d' \mid q, d)$.

This joint formulation provides a principled proxy for utility-aligned rewriting.
By the definition of conditional probability, the posterior distribution of $d'$ given an answer $a$ satisfies: 
\begin{equation}
\label{eq:posterior}
P_{\phi}(d' \mid q, d, a) = \frac{P_{\phi}(a, d' \mid q, d)}{P_{\phi}(a \mid q, d)} \propto P_{\phi}(a, d' \mid q, d).
\end{equation}
Sampling $d'$ from the LLM’s joint reasoning traces approximates sampling from this posterior, yielding rewrites that are explicitly grounded in answer generation.
As a result, rewriting supervision derived from joint trajectories is naturally aligned with document utility.

While this formulation specifies what constitutes utility-aligned rewriting, it does not yet specify how to reliably obtain such supervision or how to distinguish high-utility rewrites from low-utility ones.
We address these challenges by distilling joint process supervision from LLMs in an identifiable manner (\S\ref{subsec:inference_scaling}) and by selecting supervision signals based on measured utility improvement for a downstream generator (\S\ref{subsec:utility_supervision}).

\subsection{Scaled Process Supervision}
\label{subsec:inference_scaling}
Figure~\ref{fig:ours_overview}(a) illustrates our data collection process, where for each query--document set $(q, D)$, we collect joint supervision traces $P(a, d' \mid q, d)$ by prompting the LLM to rewrite and generate the answer within a single autoregressive trajectory.

\paragraph{Empirical observation of joint generation}
We observe that sufficiently capable LLMs naturally generate such a joint structure in their reasoning traces.
In manual inspection of reasoning traces from multiple LLMs~\footnote{From chain-of-thought prompting~\cite{wei2022chain}, 10 queries per model on Llama-3.1-\{70b, 405b\}-Instruct, claude-3-5-sonnet, Deepseek-r1, Mistral-\{large,large2\}, reka-flash, and snowflake-arctic. The examples and more detailed results are in Appendix~\ref{sec:app:llm_reasoning_process}.}, we find that rewritten documents frequently appear as intermediate steps for the answer generation, even without explicit prompting to do so, as shown in examples in Table~\ref{tab:rewriting_examples}.
This empirical evidence suggests that large LLMs implicitly generate rewritten documents and answers in a single autoregressive process, allowing us to effectively sample document $d'$ from a joint distribution $P_{\mathrm{LLM}}(a,d' \mid q, d)$.

\paragraph{Disentangling document-level utility via scaled observation}
A naive way to collect joint rewrite--answer traces is to prompt the LLM with the full retrieved document set $D$ and let it rewrite all documents before producing an answer.
However, when multiple rewritten documents $\{d'_1, \dots, d'_{|D|}\}$ are generated in a single trajectory and followed by a correct answer, it becomes difficult to disentangle which rewritten document actually contributed to the answer generation.
For instance, a correct answer may arise because one document was rewritten effectively, while others were rewritten poorly or not at all.

To resolve the challenge, we leverage inference scaling as a form of controlled observation, which scales the generation process to the document level to disentangle the supervision signals.
The scaling is depicted with the blue color in Figure~\ref{fig:ours_overview}(a).
Concretely, for each query--document set $(q, D)$, we perform $|D|$ LLM calls, each targeting a single document $d_i$ for rewriting, conditioned on the query and the remaining documents $D \setminus \{d_i\}$ as context. From scaled runs, it yields a set of isolated supervision traces $\{(d_i, d_i', a_i)\}_{i=1}^{|D|}$.
This strategy preserves the benefits of a global context that captures inter-document relations~\cite{burges2010ranknet}, while ensuring that the specific utility of the target document is explicitly disentangled.
Implementation details and prompts are provided in Appendix~\ref{sec:app:training_prompts}.

\subsection{Utility as Soft Label}
\label{subsec:utility_supervision}
To address the coarseness of answer-level supervision, we construct training signals based on \textbf{utility improvement} measured with respect to a downstream generator.
Rather than labeling rewritten documents by hard signal, we quantify soft utility gain to guide preference tuning.

\paragraph{Data curation}
For a given query $q$, retrieved document $d$, and its rewritten counterpart $d'$, we evaluate the downstream generator $\phi$ by computing the average log-likelihood of the gold answer $a^*$:
\begin{equation}
\begin{split}
    \ell(d) &= \mathbb{E}_{t} \left[ \log P_{\phi}(a^*_t \mid q, d, a^*_{<t}) \right], \\
    \ell(d') &= \mathbb{E}_{t} \left[ \log P_{\phi}(a^*_t \mid q, d', a^*_{<t}) \right]. \\
\end{split}
\end{equation}
We define the utility improvement of a rewritten document as $\Delta \ell = \ell(d') - \ell(d)$.
A rewritten document is considered utility-improving if it increases the generator’s likelihood of the correct answer.

With this utility improvement, we construct supervision signals by filtering rewritten documents.
Specifically, as depicted in Figure~\ref{fig:ours_overview}(b), rewritten documents with $\Delta \ell > 0$ or with absolute likelihood $\ell(d')$ exceeding a minimum threshold $\tau$ are treated as positive examples (\emph{win}), while the remaining documents are treated as negative examples (\emph{lose}).

This criterion ensures that training focuses on rewritten documents that measurably improve the generator’s answer distribution, rather than merely yielding correct answers.

\paragraph{Training}
We use the resulting supervision signals to train the student rewriter following standard practice.
The training procedure consists of supervised fine-tuning (SFT) followed by direct preference optimization (DPO), as commonly adopted in prior works.

For SFT, we train the rewriter to generate rewritten documents conditioned on the query and retrieved document set, using only utility-improving (\emph{win}) examples.
Concretely, given a query $q$, retrieved documents $D$, and a utility-improving rewrite $D'_w$, we maximize the log-likelihood of the rewrite:
\begin{equation}
    \mathcal{L}_{\text{SFT}}(\theta) = \mathbb{E}_{(q,D,D'_w)} \left[ \log P_{\theta}(D'_w \mid q, D) \right]
\end{equation}
where $\theta$ denotes the parameters of the student rewriter.
This stage initializes the model to produce high-quality rewrites aligned with utility-improving behavior.

Starting from the SFT-trained model, we further apply direct preference optimization.
For each query-documents pair, we construct preference pairs $(D'_w, D'_l)$ from utility-improving (\emph{win}) and non-improving (\emph{lose}) rewrites.
Following standard DPO formulations, we optimize the rewriter to assign higher likelihood to the preferred rewrite:
\begin{equation}
\begin{split}
    \mathcal{L}_{\text{DPO}}(\theta) = - \mathbb{E}_{(q,D,D'_w,D'_l)} [ &\log \sigma ( r_{\theta}(q,D,D'_w) \\
    &- r_{\theta}(q,D,D'_l) )]
\end{split}
\end{equation}
where $\sigma(\cdot)$ denotes the sigmoid function.
Following prior work~\cite{rafailov2023direct}, we simplify $r_{\theta} = \beta \log P_{\theta}(D'|q,D) - \beta \log P_{\theta_{\textrm{SFT}}}(D'|q,D)$, where $\theta_{\textrm{SFT}}$ denotes SFT-trained parameters.

\section{Experiments}
\begin{table*}[t]
\centering
\scalebox{0.87}{
    \def\arraystretch{1}
    \begin{tabular}{cccccccccccc|cc}
    \toprule
    \multirow{2}{*}{\textbf{Method}} & \multirow{2}{*}{\makecell{\textbf{True}\\\textbf{Utility}}} & \multirow{2}{*}{\textbf{Distill}} & \multirow{2}{*}{\makecell{\textbf{Module}\\\textbf{Size}}} & \multicolumn{2}{c}{\textbf{AmbigQA}} & \multicolumn{2}{c}{\textbf{HotpotQA}} & \multicolumn{2}{c}{\textbf{2WIKI}} & \multicolumn{2}{c}{\textbf{MuSiQue}} & \multicolumn{2}{|c}{\textbf{Average}} \\
    \cmidrule(lr){5-6}\cmidrule(lr){7-8}\cmidrule(lr){9-10}\cmidrule(lr){11-12}\cmidrule(lr){13-14}
    & & & & EM & F1 & EM & F1 & EM & F1 & EM & F1 & EM & F1 \\
    \midrule
    \multicolumn{1}{l}{Naive} & - & - & - & 52.8 & 64.4 & 47.2 & 63.8 & 38.4 & 52.0 & 12.1 & 20.2 & 37.6 & 50.1 \\
    \multicolumn{1}{l}{BGE-Reranker} & \xmark & \xmark & 560M & 52.8 & 64.0 & 48.7 & 65.5 & 38.7 & 53.4 & 13.0 & 20.9 & 38.3 & 51.0  \\
    \multicolumn{1}{l}{RankZephyr} & \xmark & \cmark & 7B & 52.6 & 64.0 & 49.1 & 65.8 & 39.5 & 53.9 & 12.7 & 20.5 & 38.5 & 51.1 \\
    \multicolumn{1}{l}{Rank-K} & \xmark & \cmark & 32B & 53.5 & 64.8 & 48.8 & 65.5 & 40.0 & 54.3 & 12.6 & 20.4 & 38.7 & 51.3 \\
    \multicolumn{1}{l}{Provence} & \tmark & \cmark & 430M & 52.3 & 64.7 & 43.4 & 58.1 & 35.1 & 45.1 & 12.5 & 19.6 & 35.8 & 46.9 \\
    \multicolumn{1}{l}{ComPACT} & \xmark & \cmark & 7B & 24.0 & 30.2 & 47.8 & 63.7 & 43.8 & 53.3 & 10.6 & 16.7 & 31.6 & 41.0 \\
    \multicolumn{1}{l}{Refiner} & \tmark & \cmark & 7B & 31.1 & 37.6 & 47.5 & 63.2 & 43.6 & 53.3 & 12.0 & 18.7 & 33.6 & 43.2 \\
    \multicolumn{1}{l}{EXIT} & \tmark & \xmark & 2B & 49.2 & 60.0 & 38.5 & 51.5 & 32.7 & 41.1 & 8.5 & 14.4 & 32.2 & 41.8 \\
    \midrule
    \multicolumn{1}{l}{\ours} & \cmark & \cmark & 3B & \textbf{54.5} & \textbf{67.2} & \textbf{49.4} & \textbf{68.2} & \textbf{54.4} & \textbf{65.7} & \textbf{16.0} & \textbf{26.3} & \textbf{43.6} & \textbf{56.9 } \\
    
    \bottomrule
    \end{tabular}
}
\caption{Performance comparison on AmbigQA, HotpotQA, 2Wiki, and MuSiQue using Llama-3.1-8B-Instruct. Results are reported with EM and F1, with the last two columns showing the average EM and F1 across the datasets. The best performance for each target dataset is marked in bold.}
\vspace{-5mm}
\label{tab:main_results_multihop_8b}
\end{table*}

\begin{table}[t]
\centering
\scalebox{0.82}{
    \def\arraystretch{1}
    \begin{tabular}{ccc|c}
        \toprule
        \multirow{1}{*}{\textbf{Method}} & \textbf{MS MARCO} & \textbf{CRAG} & \textbf{Average} \\
        \midrule
        \multicolumn{1}{l}{Naive}          & 65.5 \small{$\pm$ 0.3} & 34.1 \small{$\pm$ 0.3} & 49.8 \\
        \multicolumn{1}{l}{BGE-Reranker}   & 65.0 \small{$\pm$ 0.7} & 34.2 \small{$\pm$ 0.3} & 49.6 \\
        \multicolumn{1}{l}{RankZephyr}     & 66.1 \small{$\pm$ 0.5} & 34.7 \small{$\pm$ 0.2} & 50.4 \\
        \multicolumn{1}{l}{Rank-K}         & 67.5 \small{$\pm$ 0.4} & 34.3 \small{$\pm$ 0.2} & 50.9 \\
        \multicolumn{1}{l}{Provence}       & 65.5 \small{$\pm$ 0.2} & 32.9 \small{$\pm$ 0.2} & 49.2 \\
        \multicolumn{1}{l}{ComPACT}        & 55.4 \small{$\pm$ 0.6} & 29.7 \small{$\pm$ 0.3} & 42.6 \\
        \multicolumn{1}{l}{Refiner}        & 65.3 \small{$\pm$ 0.2} & 29.3 \small{$\pm$ 0.1} & 47.3 \\
        \multicolumn{1}{l}{EXIT}           & 57.0 \small{$\pm$ 0.2} & 31.5 \small{$\pm$ 0.2} & 44.3 \\
        \midrule
        \multicolumn{1}{l}{\ours}          & \textbf{68.1} \small{$\pm$ 0.1} & \textbf{35.8} \small{$\pm$ 0.3} & \textbf{52.7} \\
        \bottomrule
    \end{tabular}
}
\caption{Performance comparison on MS MARCO and CRAG using Llama-3.1-8B-Instruct. Results are reported with ACC using LLM (Llama-3.1-405B) evaluation. The last column shows the average ACC across the two datasets. The best performance for each target dataset is marked in bold.}
\label{tab:main_results_msmarco_crag_8b}
\vspace{-5mm}
\end{table}

\subsection{Experimental Setting}
In this section, we describe the datasets, metrics, and implementation details in our experiments.

\paragraph{Datasets and Metrics}
Following evaluation conventions~\cite{chirkova2025provence,yoon2024compact}, we evaluate \ours on both Wikipedia-based and web-scale question answering(QA) benchmarks to assess utility-oriented rewriting under different retrieval and reasoning conditions.

For wikipedia-based corpus, we consider multi-hop and ambiguity-focused QA datasets:
\textbf{HotpotQA}~\cite{yang2018hotpotqa},
\textbf{2WikiMultihopQA (2WIKI)}~\cite{ho2020constructing},
\textbf{MuSiQue}~\cite{trivedi2022musique}, and
\textbf{AmbigQA}~\cite{min-etal-2020-ambigqa}.
These datasets require reasoning over multiple pieces of evidence or resolving ambiguous queries by conditioning on appropriate contextual evidence.
We report Exact Match (EM) and F1 scores, following standard QA evaluation~\cite{hwang2024exit, yoon2024compact}.

For web-scaled corpus to assess practical effectiveness under web-scaled retrieval conditions, we additionally use web-based QA benchmarks.
We include \textbf{MS MARCO}~\cite{DBLP:journals/corr/NguyenRSGTMD16} for single-hop QA, evaluated on a randomly sampled subset of 500 queries following prior work~\cite{lee-etal-2025-inference,back-etal-2021-learning}.
For more challenging reasoning scenarios beyond extraction, we evaluate on \textbf{CRAG}~\cite{yang2024crag}, which covers diverse query types requiring synthesis and reasoning over web documents.
For these datasets, we report Accuracy (ACC) using an LLM-based evaluation protocol with Llama-3.1-405B, following the dataset’s evaluation setup~\cite{yang2024crag}\footnote{We conducted a reliability analysis of LLM-as-a-judge evaluation in Appendix~\ref{sec:app:analysis_llm_evaluation}.}.

Following~\citet{li2024refiner}, we use the retrieval contexts provided by each dataset and fix the number of retrieved documents to top-10.
We note that datasets such as MuSiQue and CRAG often do not include gold documents in the retrieved set, making them well-suited for evaluating utility-oriented rewriting beyond extraction.

\paragraph{Baselines}
For reranking-based baselines, we include \textbf{BGE-Reranker-Large}~\cite{bge_embedding}, a cross-encoder pointwise reranker that scores documents independently, and \textbf{RankZephyr}~\cite{pradeep2023rankzephyr}, a listwise reranker distilled from GPT-4 that jointly models document interactions.
We further include \textbf{Rank-K}~\cite{yang2025rank}, a recent strong reranker that incorporates explicit reasoning signals.
These methods represent strong reranking-based bridges that improve utility through document ordering without modifying document content.

For summarization-based baselines, we compare against \textbf{ComPACT}~\cite{yoon2024compact} and \textbf{Refiner}~\cite{li2024refiner}, which distill LLM knowledge to generate query-focused abstractive summaries, and \textbf{EXIT}~\cite{hwang2024exit}, which improves generation by pruning irrelevant document content.
These approaches explicitly rewrite document content to better align with the query.

For unified baseline, we include \textbf{Provence}~\cite{chirkova2025provence}, which unifies reranking and summarization into a single bridge model.

\paragraph{Implementation details}
We use Llama-3.3-70B-Instruct~\cite{llama3modelcard} as the teacher model to generate joint rewrite--answer trajectories.
For utility-based preference, we set $\kappa=4.0$ and the utility threshold $\tau=0.05$ to strictly divide the preference, following CoRAG~\cite{wang2025corag}.
To prevent answer leakage, we strictly filter out queries for which the teacher can generate the correct answer without conditioning on retrieved documents.

From the generated traces, we construct two datasets from MS MARCO:
14k utility-improving rewrite examples for SFT, and 8k preference pairs for DPO.
These datasets are derived by measuring the utility improvement of rewritten documents with respect to Llama-3.2-3B-Instruct as the downstream generator.

We train the student rewriter with Llama-3.2-3B-Instruct as the backbone.
SFT is conducted using DeepSpeed ZeRO-3~\cite{rasley2020deepspeed} on four NVIDIA A6000 GPUs for three epochs, followed by DPO for 500 steps with $\beta=0.1$.
Inference is performed with vLLM~\cite{kwon2023efficient}, using a temperature of 0 for reproducibility.

\subsection{Results}

\paragraph{Research Questions} 
To evaluate the effectiveness of our approach, we address the following
research questions: 
\begin{itemize}
    \item RQ 1: Does \ours enhance the performance of the models in QA tasks?
    \item RQ 2: Does \ours generalize to diverse generators?
    \item RQ 3: Does each component of \ours contribute to overall effectiveness? 
\end{itemize}

\begin{figure}[t]
    \centering
    \includegraphics[width=1\linewidth]{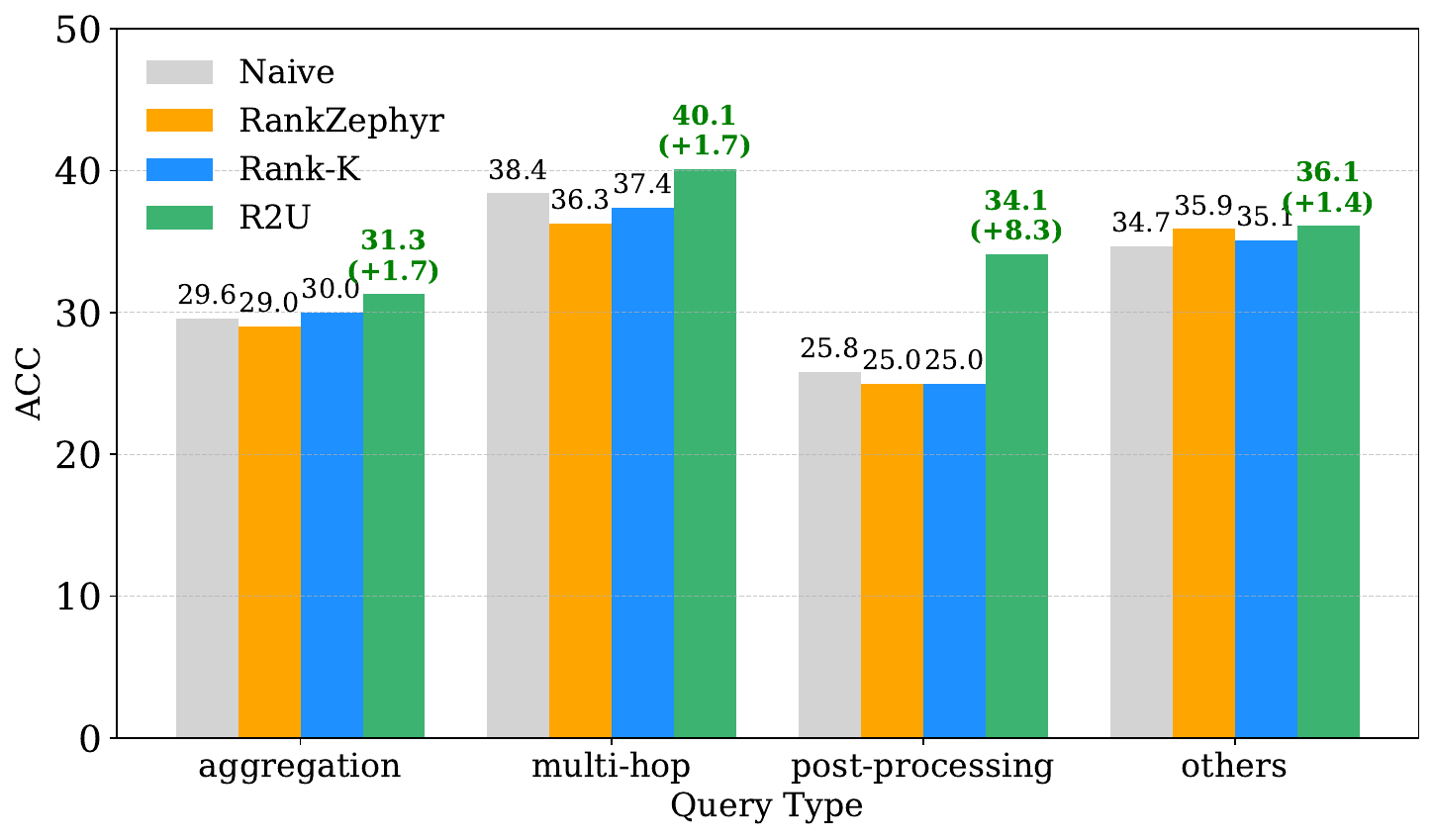}
    \caption{Comparison of ACC across various query types in CRAG. The value in parentheses indicates the difference relative to Naive RAG.}
    \label{fig:crag_analysis}
    \vspace{-5mm}
\end{figure}

\subsection{RQ 1: Does \ours enhance the performance of the models in QA tasks?}

\paragraph{Overall Performance and True Utility}
Table~\ref{tab:main_results_multihop_8b} and~\ref{tab:main_results_msmarco_crag_8b} highlight the dominant performance of \ours compared to various baselines in question answering tasks.
We separate the results to focus on different evaluation aspects:
Table~\ref{tab:main_results_multihop_8b} assesses disambiguation or multi-hop reasoning ability on the Wikipedia corpus,
and Table~\ref{tab:main_results_msmarco_crag_8b} evaluates performance on the web corpus to validate that our gains in document utility generalize to practical settings.

As shown in the column named ``True Utility'' in Table~\ref{tab:main_results_multihop_8b}, baselines are marked with \xmark~if they use a utility function without considering the answer (i.e., $P_{\theta}(d' \mid q, d)$ only) and with \tmark~if they incorporate the answer supervision (Eq.~\ref{eq:result_sup}).
Overall, \ours achieves the highest scores, demonstrating particularly strong performance on disambiguation and multi-hop queries, where cross-document reasoning is required to answer the query.
Compared to baselines, \ours outperforms NaiveRAG by +6.8\%pt and the strongest competitor by +5.6\%pt in terms of average F1, supporting the hypothesis that aligning document utility with answer generation effectively guides reasoning over multiple sources.
Table~\ref{tab:main_results_msmarco_crag_8b} further shows that \ours maintains superior performance on the web corpus, including MS MARCO and CRAG.
This confirms that the benefits of our true utility function extend beyond structured Wikipedia datasets to more diverse, open-domain web content.

\paragraph{Comparison with Distilled Baselines}
While baselines such as RankZephyr~\cite{pradeep2023rankzephyr} and Refiner~\cite{li2024refiner} also leverage knowledge distillation from large language models for document rewriting, their performance remains lower than that of \ours.
This indicates that the observed gains primarily stem from aligning document utility, rather than from distillation alone.
Notably, \ours achieves superior performance with a comparatively small 3B module, showing that our approach enables effective distillation even with limited model capacity.

\paragraph{Performance on Complex Query Types}
To further investigate the role of document utility in complex query resolution, we categorize CRAG queries into four types: aggregation, multi-hop, post-processing-heavy, and others~\cite{yang2024crag}.
This categorization allows us to disentangle the varying levels of document interaction and reasoning required by different query types. We compare \ours against RankZephyr and Rank-K, which achieve the strongest overall performance among existing baselines on CRAG.

As shown in Figure~\ref{fig:crag_analysis}, \ours attains the highest performance across all complex query types. 
While Rank-K also slightly improves performance on aggregation queries, this gain comes at the cost of accuracy on other types. In contrast, \ours demonstrates consistently superior performance, with particularly pronounced gains in scenarios demanding higher document utility.
Specifically, \ours outperforms alternatives on aggregation and multi-hop queries, where the effective integration of multiple documents is crucial.
The improvement is even more substantial for post-processing-heavy queries, which require both multi-document reasoning and significant post-processing.

These findings highlight that by enhancing the true utility of retrieved documents through the bridge module, \ours enables the generator to more effectively leverage relevant evidence, outperforming baselines.

\begin{figure}[t]
    \centering
    \includegraphics[width=1\linewidth]{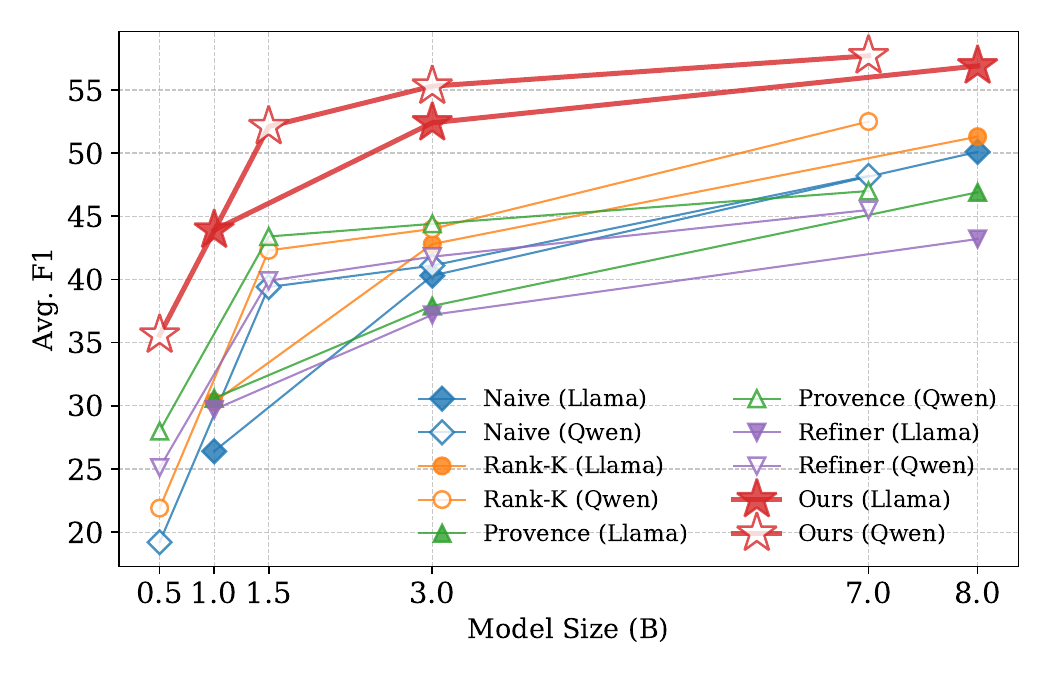}
    \caption{Comparison of average F1 scores across various model sizes using Llama (filled markers) and Qwen (hollow markers). 
    }    
    \label{fig:f1_vs_modelsize}
    \vspace{-5mm}
\end{figure}


\begin{table*}[t]
\centering
\scalebox{0.8}{
    \def\arraystretch{1}
    \begin{tabular}{cccccccccccc}
    \toprule
    \multirow{2}{*}{\textbf{Method}} & \multicolumn{1}{c}{\textbf{MS MARCO}} & \multicolumn{1}{c}{\textbf{CRAG}}  & \multicolumn{1}{c}{\textbf{AmbigQA}} & \multicolumn{1}{c}{\textbf{HotpotQA}} & \multicolumn{1}{c}{\textbf{2WIKI}} & \multicolumn{1}{c}{\textbf{MuSiQue}}  \\
    \cmidrule(lr){2-2}\cmidrule(lr){3-3}\cmidrule(lr){4-4}\cmidrule(lr){5-5}\cmidrule(lr){6-6}\cmidrule(lr){7-7}
    & ACC & ACC  & F1 & F1 & F1 & F1 \\
    \midrule
    \multicolumn{1}{l}{Naive RAG}     & 70.5 & 34.0 & 64.4 & 64.0 & 52.1 & 19.9 \\
    \multicolumn{1}{l}{ComPACT (70B)} &  \textbf{70.7}  & 35.6 & 69.8 & 72.9 & 66.3 & 24.7 \\ \midrule
    \multicolumn{1}{l}{\ours (70B)} & \textbf{70.7} & \textbf{37.2}  & \textbf{71.8} & \textbf{73.7} & \textbf{67.8} & \textbf{37.1} \\
    
    \bottomrule
    \end{tabular}
}
\caption{Comparison of \ours with ComPACT using Llama-3.3-70B-Instruct as a rewriter and Llama-3.1-8B-Instruct as a generator. The best performance for each target dataset is marked in bold.}
\label{tab:ablation_70b_compact}
\end{table*}

\begin{table*}[t]
\centering
\scalebox{0.90}{
    \def\arraystretch{1}
    \begin{tabular}{ccccccccc|cc}
    \toprule
    \multirow{2}{*}{\textbf{Method}} & \multicolumn{2}{c}{\textbf{AmbigQA}} & \multicolumn{2}{c}{\textbf{HotpotQA}} & \multicolumn{2}{c}{\textbf{2WIKI}} & \multicolumn{2}{c}{\textbf{MuSiQue}} & \multicolumn{2}{|c}{\textbf{Average}} \\
    \cmidrule(lr){2-3}\cmidrule(lr){4-5}\cmidrule(lr){6-7}\cmidrule(lr){8-9}\cmidrule(lr){10-11}
    & EM & F1 & EM & F1 & EM & F1 & EM & F1 & EM & F1 \\
    \midrule
    \multicolumn{1}{l}{\ours} & \textbf{54.5} & \textbf{67.2} & \textbf{49.4} & \textbf{68.2} & \textbf{54.4} & \textbf{65.7} & \textbf{16.0} & \textbf{26.3} & \textbf{43.6} & \textbf{56.9 } \\
    \midrule
    \multicolumn{1}{l}{w/o DPO} &  51.9 & 64.7  & 48.7 & 66.5 & 49.6 & 61.1  & 14.3 & 22.5  & 41.1 & 53.7 \\
    \multicolumn{1}{l}{w/o DPO and soft label} & 53.6 & 66.4 & 48.7 & 66.8 & 45.8 & 59.4 & 12.7 & 20.9 & 40.2 & 53.4 \\
    \multicolumn{1}{l}{w/o DPO and scaled obs.} & 44.3 & 53.1 & 38.4 & 51.8 & 28.6 & 36.5 & 9.1 & 15.2 & 30.1 & 39.2 \\
    
    
    \bottomrule
    \end{tabular}
}
\caption{Ablation study. The best performance for each target dataset is marked in bold.}
\vspace{-5mm}
\label{tab:ablation}
\end{table*}

\subsection{RQ 2: Does \ours generalize to diverse generators?}

To address RQ2, we evaluate \ours with various generators, against the best-performing module of each type based on F1 score: Rank-K for reranking, Refiner for summarization, and Provence for combining both.
We use Llama (Llama-3.2-\{1B,3B\}-Instruct, Llama-3.1-8B-Instruct) and Qwen (Qwen2.5-\{0.5B,1B,3B,7B\}-Instruct) Family~\cite{bai2023qwen} as generators.

The results in Figure~\ref{fig:f1_vs_modelsize} demonstrate the effectiveness and generalizability of our method over different sizes and types of generators.
Notably, \ours enables Llama-1B and Qwen-1.5B to outperform their larger counterparts (Llama-3B and Qwen-3B) with other methods.
These results highlight that smaller models that lack strong reasoning capabilities often gain more from the effectiveness of \ours.
The full results are available in Table~\ref{tab:multihop_results_llama1b}-\ref{tab:multihop_results_qwen7b} in Appendix~\ref{sec:app:per_model_size}.

\begin{figure}[t]
    \centering
    \includegraphics[width=0.8\linewidth]{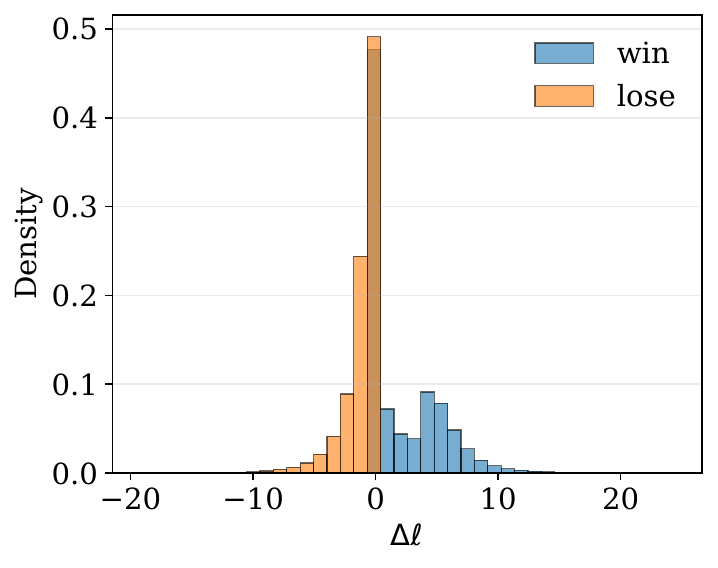}
    \caption{Distribution of utility improvement ($\Delta \ell$) for win and lose rewrites.}
    \label{fig:delta_hist}
    \vspace{-5mm}
\end{figure}
\subsection{RQ 3: Does each component of \ours contribute to overall effectiveness?}
\label{subsec:rq3}
We conduct an ablation study to examine the contribution of each component of R2U:
(i) joint process supervision,
(ii) utility-based soft label, and
(iii) scaled observation for disentangling document-level utility.

\paragraph{Effect of joint process supervision.}
Table~\ref{tab:ablation_70b_compact} compares our joint rewrite–answer supervision against a disjoint rewriting baseline adapted from ComPACT, where documents are rewritten independently of answer generation.
Using the same large rewriter (Llama-3.3-70B-Instruct), R2U consistently outperforms the disjoint baseline across all datasets.
This confirms that coupling rewriting with answer generation is crucial for capturing document utility aligned with downstream reasoning, rather than merely producing query-relevant summaries.

\paragraph{Effect of training objectives (SFT vs. DPO).}
Table~\ref{tab:ablation} shows that removing DPO (w/o DPO) leads to a clear performance drop, although the SFT-only variant still outperforms most baselines.
This indicates that SFT on utility-improving rewrites already provides strong gains, while DPO further refines the model by sharpening distinctions between high- and low-utility rewrites.
To enable a more controlled analysis, we first ablate DPO and conduct the remaining ablation studies on the SFT-trained model, thereby isolating the effect of each component.

\paragraph{Effect of utility-based soft label.}
To isolate the role of utility-based supervision selection, we compare R2U with a naive variant that trains on all rewrites generated by the teacher model without soft labeling (w/o soft label).
In R2U (w/o DPO), rewrites with negative utility improvement ($\Delta \ell < 0$) are explicitly excluded from training, preventing counterproductive supervision, as shown in the left of Figure~\ref{fig:delta_hist}.
In contrast, the w/o soft label variant indiscriminately learns from rewrites, including those that reduce the downstream generator’s likelihood of the correct answer.
Notably, in Table~\ref{tab:ablation}, this difference leads to weaker performance, since rewrites with $\Delta \ell < 0$ are treated as positive training signals in w/o soft label.
These results indicate that soft label based on explicit utility improvement is critical for learning utility-aligned rewriting behavior.

\paragraph{Effect of scaled observation.}
Finally, removing scaled observation (w/o scaled obs.) causes the largest performance degradation. This variant collapses to naive joint rewriting, where document-level contributions are entangled within a single trajectory. The sharp drop in Table~\ref{tab:ablation} demonstrates that isolating each document’s contribution via scaled, controlled observations is essential for learning reliable document-level utility signals.

\section{Conclusion}

In this work, we address the gap between 
retriever and generator in RAG. 
We propose to observe reasoning process as process supervision, then distill to a smaller model for practical deployment. Moreover, \ours enables more effective use of retrieved information for answer generation. Empirical results on diverse datasets confirm that our rewritten documents significantly improve performance over strong baselines.
\section{Limitation}
While \ours shows significant gains, generating process-supervised data relies on costly LLM reasoning traces. Due to this expense, we were unable to conduct data generation with larger teacher models beyond 70B. Moreover, our evaluation is focused on QA tasks, leaving broader RAG applications for future work.
As with other distillation-based approaches, our method inherently depends on the quality of the teacher model.
Though our utility-improvement data curation and potential shortcut filtering are designed to mitigate spurious supervision, fully eliminating all non-causal patterns inherited from the teacher remains challenging.

\bibliography{custom}
\appendix
\newpage
\newpage
\section*{Appendix}

\section{Details and Examples of LLM's Reasoning Process}
\label{sec:app:llm_reasoning_process}

We conducted a systematic inspection across 8 model families with 10 MS MARCO examples, finding that this rewriting behavior occurred in 95\% of cases.
We elicit reasoning traces with the following prompt:
\begin{quote}\small
\texttt{Think step by step to use the provided documents to answer a user's question.}\\
\texttt{Question: \dots}\\
\texttt{Documents: \\
Document 1: \dots \quad \\
Document 2: \dots \quad \\
\dots}
\end{quote}
A step is annotated as a rewrite when the trace explicitly cites one or more source documents (e.g., ``from Document 1, 2, ...'' or ``Document 1 supports \dots'') A case satisfies the prerequisite if at least one such rewrite appears before the final answer. Table~\ref{tab:rewriting_examples} presents representative cases, illustrating how robust LLMs produce $d'$ tailored to the target answer under its reasoning trace.

We expanded this manual inspection to quantify how frequently LLMs naturally perform rewriting across three distinct datasets, where we sampled 50 examples and evaluated them manually. As shown in Table~\ref{tab:inherent_rewriting_multidata}, LLMs exhibit this behavior in nearly 100\% of cases across all tasks.

\begin{table}[h]
\centering
\scalebox{0.87}{
\begin{tabular}{l|ccc}
\toprule
Model & CRAG & HotpotQA & MS MARCO \\
\midrule
Llama-3.1-70B    & 98\%  & 100\% & 100\% \\
Mistral-Large2   & 98\%  & 98\%  & 98\%  \\
\bottomrule
\end{tabular}
}
\caption{Percentage of inherent rewriting across datasets.}
\label{tab:inherent_rewriting_multidata}
\end{table}

\begin{table*}[t]
    \centering
    \begin{tabular}{p{\linewidth}}
        \thickhline
         \emph{Question:} does xpress bet charge to deposit money in your account
        \\\hline
        \emph{Claude-3.5-sonnet:} \newline
        Let me analyze this step by step: \newline
        First, I'll look for specific information about deposit fees in the documents.\newline
        From \textbf{Document 7}:\newline
        ``They also give you the ability to wager for free online, make free deposits, and have free withdrawals.''\newline
        From \textbf{Document 9}:\newline
        ``No membership, monthly, or per-wager fees''\newline
        However, there is one exception noted in \textbf{Document 2:}\newline
        When using certain payment methods like PayNearMe, there may be a convenience fee (normally \$3.99) \newline
        Looking at deposit methods available (from Documents 4 and 9): ... \\ \hline
        
        \emph{Deepseek-r1}: \newline
        <think> Okay, let me try to figure out if Xpressbet charges a fee for depositing money into an account. The user is asking specifically about deposit fees. \newline
        First, I'll go through each document provided to see if any mention of fees is made. Let's start with \textbf{Document 1}. It talks about the minimum deposit amounts (\$10 in most states, \$25 in Massachusetts) but doesn't mention any fees. It does mention a bonus when a friend deposits and wagers, but that's not related to fees. \newline
        \textbf{Document 2} is more relevant. The first step says, ``Tell the cashier how much you want to deposit... they will collect your funds plus the convenience fee (normally \$3.99).'' So here, there's a convenience fee of \$3.99 when depositing via a cashier using PayNearMe, I think. That's a direct mention of a fee for deposit. \newline
        \textbf{Document 3} repeats the PayNearMe steps but doesn't mention the fee again. \textbf{Document 4} lists deposit methods like Xpressfund, Green Dot, PayNearMe, etc., but doesn't state fees. \textbf{Document 5} is similar to 2 and 3, again about PayNearMe steps without mentioning the fee here. \newline
        \textbf{Document 6} is about mobile PayCode setup; again no fee mentioned. \textbf{Document 7} is really important. It says, ''make free deposits, and have free withdrawals.'' So this directly states that deposits are free. But wait, there's a conflict here with \textbf{Document 2} which mentions a \$3.99 fee. Maybe the fee depends on the deposit method? \newline
        Looking at \textbf{Document 7} again: ``They also give you the ability to wager for free online, make free deposits, and have free withdrawals.'' So according to this, deposits are free. But Document 2's convenience fee is for a specific method, PayNearMe via cashier. So maybe some methods have fees, others don't. \newline
        \textbf{Document 9} repeats that there are no per-wager fees and lists deposit methods again without mentioning fees, except for the possible ones in Document 2. Document 8 and 10 talk about other services (U.S. Bank Mobile app, Western Union) which might not be directly related to Xpressbet's own fees. \newline
        So putting it all together: ... \\ \hline
        
        \emph{llama-3.1-405b:} \newline
        To answer the user's question, we need to find information about Xpressbet's deposit fees. Let's analyze the documents: \newline 
        \textbf{Document 2} mentions a convenience fee of \$3.99 when depositing funds into an Xpressbet account using a cashier. \newline
        \textbf{Document 7} states that Xpressbet does not charge per wager fees, and also mentions that deposits are free. \newline
        \textbf{Document 9} also mentions that there are no membership, monthly, or per-wager fees, and lists various deposit methods, but does not mention any fees associated with depositing funds. \newline
        Based on the information provided, it appears that ... \\ \hline
        \thickhline
    \end{tabular}
    \caption{An example from the MS MARCO development set where LLMs naturally perform rewriting during their reasoning process. Models were provided with all 10 passages to generate the answer.}
    \label{tab:rewriting_examples}
\end{table*}

\section{Prompts for \ours}
\label{sec:app:training_prompts}
Table~\ref{tab:training_prompt_template} provides the full CoT prompt template for generating scaled process supervision. 
We design a revised CoT prompt template to observe the desired document distribution from the reasoning process.
To ensure better supervision quality, we have made several design choices:
(1) Separate the document rewriting step from the reasoning and answering step, and discard the latter step to prevent leakage. (2) Restrict rewrites to retrieved content. (3) Encourage cross-document interaction. (4) Allow skipping irrelevant documents.

For each query with a set of retrieved documents $D$ (typically $|D|=10$), we executed $|D|$ separate generation passes. In the $i$-th pass (where $i \in \{1, \dots, |D|\}$), we designated the current document $d_i$ as the \{document\} and the remaining set $D \setminus \{d_i\}$ as the \{other documents\}. Every document is rewritten exactly once as the primary target while being conditioned on the global context of all other documents. We then aggregated these individual traces to form the final supervision set $\{(d_i, d'_i, a_i)\}_{i=1}^{|D|}$ for the query.

Table~\ref{tab:r2u_prompt_template} presents the \ours prompt template for document rewriting during both training and inference.
The template takes a query and the full set of retrieved documents as input and rewrites them in a single forward pass.

\begin{table*}[t]
    \centering
    \begin{tabular}{p{\linewidth}}
        \thickhline
         \textbf{CoT prompt template for generating scaled process supervision}
        \\\hline
You are given a user query, a set of other documents, and one target document. Your task is to answer the query based solely on the target document, while using the other documents to help clarify and improve the target document. To do this, you will first rewrite the target document to make it more accurate, complete, and easier to reason with in the context of the query.\newline
Step 1: Rewrite the **target document** using the other documents so that it is accurate, non-redundant, and maximally informative in light of the query.\newline
Step 2: Use the rewritten document to answer the query. Think step by step and explain your reasoning.\newline

[Step 1 Guidelines]\newline
1. Relevance Check\newline
- If parts of the target document are not relevant to the query or not logically connected to it, remove them without explanation.\newline
- Partial relevant information is acceptable. The document does not need to be sufficient to fully answer the query.\newline
- If the document does not contain any content relevant to the query, return exactly: NO REWRITE\newline

2. Integration\newline
- If information in the target document appears similar to what is in the other documents, do not overwrite or remove information—highlight subtle differences or unique nuances.\newline
- If the target document contradicts the other documents, preserve the target document’s version.\newline
- When rewriting, preserve original terminology from the target document or reuse vocabulary from the query wherever possible, rather than paraphrasing unnecessarily.\newline
- Enrich or clarify the target version using the other documents if helpful, but maintain its perspective as primary.\newline

3. Focus \& Prioritization\newline
- Emphasize information that most directly supports answering the query.\newline
- Less essential or tangential content may be shortened, but not removed if it offers a unique nuance or framing.\newline

[IMPORTANT RULES]\newline
- Do **not** answer the query directly in Step 1.\newline
- Do **not** mention the existence of other documents.\newline
- Do **not** explain your changes or reference the task.\newline
- Do **not** include statements like "target document does not provide" or "according to other documents."\newline
- Return only the rewritten version of the target document with no commentary or formatting.\newline

[Query]\newline
\textcolor{blue}{\{query\}}\newline
[Other Documents]\newline
\textcolor{blue}{\{other documents\}}\newline
[Target Document]\newline
\textcolor{blue}{\{document\}}\newline

You must follow the output format exactly as specified below.\newline
Output format:\newline
Step 1. Document Rewriting: <your rewritten version of the target document> \newline
Step 2. Answer: <your answer to the query, along with explanation based only on the rewritten document> \\ \hline
        \thickhline
    \end{tabular}
    \caption{CoT prompt template for generating scaled process supervision. It observes the desired document distribution from the reasoning process.}
    \label{tab:training_prompt_template}
\end{table*}
\begin{table*}[t]
    \centering
    \begin{tabular}{p{\linewidth}}
        \toprule
        \textbf{R2U prompt template for document rewriting} \\
        \midrule
        You are a helpful assistant. Your job is to analyze the documents below and rewrite only the parts that help clarify or refine the information in relation to the question.\\
        List each relevant document to better support answering the question. Do not include unrelated documents.\\
                    
        Question:\\
        \textcolor{blue}{\{query\}}\\
                    
        Documents:\\
        \textcolor{blue}{\{documents\}}\\
        \bottomrule
    \end{tabular}
    \caption{\ours prompt template for document rewriting.}
    \label{tab:r2u_prompt_template}
\end{table*}

\section{Details of Computational Cost}
The computational cost of LLM for R2U is incurred only once during offline supervision data construction, while its inference-time cost is identical to that of standard bridge models.

For dataset construction, we perform $14$k queries with $10$ retrieved documents each, resulting in $140$k LLM calls.
Each call produces an average of $100.84$ output tokens, yielding approximately $14$ million generated tokens in total.
In contrast, the strongest baseline, Rank-K, reports $100$k LLM calls with an average of $2{,}408.85$ output tokens per call, resulting in approximately $240$ million generated tokens.
Despite using more LLM calls, R2U requires roughly $6\times$ fewer generated tokens, as each call rewrites only a single document rather than the entire document set.

Moreover, Rank-K relies on DeepSeek-R1, a larger and more computationally expensive model, whereas R2U uses Llama-3.3-70B-Instruct as the teacher.
As a result, the total cost of constructing supervision data for R2U is substantially lower than that of Rank-K.

\section{Analysis of LLM evaluation}
\label{sec:app:analysis_llm_evaluation}
\begin{table}[t]
\centering
\scalebox{0.87}{
    \def\arraystretch{1}
    \begin{tabular}{ccc|c}
    \toprule
    \multirow{1}{*}{\textbf{Method}} & \textbf{MS MARCO} & \textbf{CRAG} & \textbf{Average} \\
    \midrule
    \multicolumn{1}{l}{Naive} & 58.6 & 34.1 & 46.4 \\
    \multicolumn{1}{l}{BGE-Reranker} & 59.0 & 34.3 & 46.7 \\
    \multicolumn{1}{l}{RankZephyr} & 59.8 & 34.2 & 47.0 \\
    \multicolumn{1}{l}{Rank-K} & 59.4 & 34.1 & 46.8 \\
    \multicolumn{1}{l}{Provence} & 58.4 & 32.4 & 45.4 \\
    \multicolumn{1}{l}{ComPACT} & 50.0 & 29.4 & 39.7 \\
    \multicolumn{1}{l}{Refiner} & 58.8 & 30.7 & 44.8 \\
    \multicolumn{1}{l}{EXIT} & 50.8 & 31.7 & 41.3 \\
    \midrule
    \multicolumn{1}{l}{\ours} & \textbf{62.4} & \textbf{35.1} & \textbf{48.8} \\
    \bottomrule
    \end{tabular}
}
\caption{Performance comparison on MS MARCO and CRAG, using Claude-3.5-Sonnet as an LLM evaluator. The best performance for each target dataset is marked in bold.}
\label{tab:main_results_msmarco_crag_8b_claudeeval}
\end{table}
To quantify the reliability of LLM evaluation, we conducted a human verification study on 100 sampled instances (50 from MS MARCO, 50 from CRAG), where three of the authors independently annotated the model outputs.
The results demonstrate that Llama-3.1-405B achieves the highest alignment with human judgment (Cohen’s Kappa: 0.960). We compared this against other strong evaluators, including Claude-3.5-Sonnet (0.900) and Mistral-Large2 (0.742). The significantly higher agreement of Llama-3.1-405B confirms it is the most reliable proxy for human evaluation in this context, justifying its usage as an evaluator.

To definitively rule out evaluator bias, we also conducted the cross-model evaluation using Claude-3.5-Sonnet as an independent judge. The results in Table~\ref{tab:main_results_msmarco_crag_8b_claudeeval} confirm that R2U consistently outperforms the baselines under the different evaluators, demonstrating that our improvements are robust and not an artifact of self-preference bias.

\section{Related Works}

\subsection{Retrieval and Generation Gap in RAG}

Retrieval modules~\cite{bm25} in RAG systems are commonly optimized using ranking metrics such as nDCG and MRR, which rely on relevance judgments based on expected human behavior~\cite{ndcg}. 
However, these metrics do not fully capture the needs of generative models consuming the retrieved context. 
For example, LLMs often exhibit positional biases distinct from humans, disproportionately focusing on tokens at the beginning and end of contexts while neglecting those in the middle~\cite{liu2024lost}. 
Furthermore, LLMs are sensitive to distracting passages~\cite{amiraz-etal-2025-distracting, cuconasu2024power}, and performance degradation from distracting passages becomes particularly severe when paired with strong retrievers~\cite{jinlong}. 
Consequently, a bridging model~\cite{ke2024bridging} is necessary to align retrieval outputs with the generative model’s requirements, thereby reconciling human-based retrieval optimization with generation needs to improve the RAG performance.

\subsection{Bridging the Gap}

Optimizing relevance in the retriever may not align with what the generation needs. Bridging models aim to mediate this gap by re-ranking or rewriting retrieved content before it is passed to a generator.

One direction is rerankers, which reorder documents to reflect usefulness to the query better. Pointwise rerankers such as BGE~\cite{bge_embedding} score each document individually, while listwise approaches like RankZephyr~\cite{pradeep2023rankzephyr} consider global ordering, often distilled from strong LLMs like GPT‑4. Recent works such as Rank‑K~\cite{yang2025rank} go further by distilling reasoning‑oriented models (e.g., DeepSeek‑R1). However, the rerankings are performed independently of the generator’s answer generation.

A second line of bridging work uses QFS, either selection-based or generation-based. Selection methods select salient segments directly from retrieved text, as EXIT~\cite{hwang2024exit}, which uses GPT‑4o pseudo‑labels to train sentence selectors but risks spurious success when reasoning shortcuts appear. Generation-based methods instead rewrite evidence with LLMs: RECOMP‑Abst~\cite{xu2023recomp} distills GPT‑3.5 summaries, while later works like CompAct~\cite{yoon2024compact} and Refiner~\cite{li2024refiner} scale to stronger (GPT-4) or multiple LLMs. Some QFS methods also filter the training dataset based on answer correctness.~\cite{hwang2024exit,yoon2024compact,chirkova2025provence}.

Our work follows the generation-based direction but differs by exploiting the reasoning traces of LLMs. Instead of only distilling summaries, we capture how the model reasons about the query to ensure document rewriting reflects the reasoning path that yields higher answerability.

\section{Performance of \ours with Different Model Sizes}
\label{sec:app:per_model_size}

Tables~\ref{tab:multihop_results_llama1b}–\ref{tab:multihop_results_qwen7b} provide the per-dataset results, including both EM and F1, corresponding to the average F1 trends shown in Figure~\ref{fig:f1_vs_modelsize}. The results consistently show that \ours outperforms the baselines across all datasets and model sizes.

\section{Ai Assistants In Research Or Writing}
We used ChatGPT to shorten the script to fit within the page limit.

\section{Use Or Create Scientific Artifacts}
We used existing scientific artifacts for research purposes, and the use of existing artifacts was consistent with their intended applications. 
\paragraph{Used Datasets}
The datasets we used are in English and are anonymized and protected from offensive content.
\begin{itemize}
    \item MS MARCO dataset~\cite{DBLP:journals/corr/NguyenRSGTMD16}: \href{ https://microsoft.github.io/msmarco/}{non-commercial research purposes only}
    \item CRAG dataset~\cite{yang2024crag}: \href{https://github.com/facebookresearch/CRAG}{CC BY-NC 4.0 license}
    \item AmbigQA dataset~\cite{min-etal-2020-ambigqa}: \href{https://nlp.cs.washington.edu/ambigqa/}{CC BY-SA 3.0 license}
    \item HotpotQA dataset~\cite{yang2018hotpotqa}: \href{https://hotpotqa.github.io/?utm_source=chatgpt.com}{CC BY-SA 4.0 license}
    \item 2WIKI~\cite{ho2020constructing}: \href{https://github.com/Alab-NII/2wikimultihop?tab=Apache-2.0-1-ov-file}{Apache 2.0 license}
    \item MusiQue~\cite{trivedi2022musique}: \href{https://github.com/StonyBrookNLP/musique}{CC BY 4.0 license}
\end{itemize}

\paragraph{Used Models}
\begin{itemize}
    \item Qwen-2.5~\cite{bai2023qwen}: \href{https://huggingface.co/Qwen/Qwen2.5-7B-Instruct}{Apache license 2.0}
    \item Llama-3~\cite{llama3modelcard}: \href{https://www.llama.com/llama3/license/}{custom commercial license}
\end{itemize}

\begin{table*}[t]
\centering
\begin{tabular}{ccccccccc|cc}
\toprule
\multirow{2}{*}{\textbf{Method}} & \multicolumn{2}{c}{\textbf{AmbigQA}} & \multicolumn{2}{c}{\textbf{HotpotQA}} & \multicolumn{2}{c}{\textbf{2WIKI}} & \multicolumn{2}{c}{\textbf{MuSiQue}} & \multicolumn{2}{|c}{\textbf{Average}} \\
 & EM & F1 & EM & F1 & EM & F1 & EM & F1 & EM & F1  \\
\midrule
\multicolumn{1}{l}{Naive} & 33.0 & 41.7 & 20.1 & 30.3 & 19.8 & 26.6 & 2.9 & 7.1 & 19.0 & 26.4 \\
\multicolumn{1}{l}{BGE-Reranker} & 32.2 & 40.6 & 24.1 & 35.6 & 20.8 & 27.8 & 3.7 & 8.0 & 20.2 & 28.0 \\
\multicolumn{1}{l}{RankZephyr} & 33.6 & 42.3 & 24.3 & 35.9 & 21.9 & 29.1 & 3.3 & 7.7 & 20.8 & 28.8 \\
\multicolumn{1}{l}{Rank-K} & 35.3 & 43.0 & 25.9 & 37.1 & 25.1 & 32.0 & 4.5 & 8.9 & 22.7 & 30.3 \\
\multicolumn{1}{l}{Provence} & 38.8 & 47.8 & 23.3 & 33.2 & 25.5 & 32.1 & 5.4 & 9.4 & 23.3 & 30.6 \\
\multicolumn{1}{l}{ComPACT} & 19.1 & 25.3 & 30.2 & 41.6 & 32.1 & 39.0 & 5.5 & 9.8 & 21.7 & 28.9 \\
\multicolumn{1}{l}{Refiner} & 22.5 & 29.2 & 29.5 & 40.6 & 30.7 & 37.4 & 6.9 & 11.6 & 22.4 & 29.7 \\
\multicolumn{1}{l}{EXIT} & 28.3 & 37.7 & 14.9 & 25.9 & 13.2 & 22.2 & 1.9 & 6.2 & 14.6 & 23.0 \\ \midrule
\multicolumn{1}{l}{\ours} & \textbf{50.1} & \textbf{60.3} & \textbf{35.3} & \textbf{49.4} & \textbf{38.6} & \textbf{47.4} & \textbf{11.4} & \textbf{18.6} & \textbf{ 33.9 } & \textbf{43.9 } \\
\bottomrule
\end{tabular}
\caption{Results on Llama3.2-1B-Instruct with EM and F1 scores.}
\label{tab:multihop_results_llama1b}
\end{table*}
\begin{table*}[t]
\centering
\begin{tabular}{ccccccccc|cc}
\toprule
\multirow{2}{*}{\textbf{Method}} & \multicolumn{2}{c}{\textbf{AmbigQA}} & \multicolumn{2}{c}{\textbf{HotpotQA}} & \multicolumn{2}{c}{\textbf{2WIKI}} & \multicolumn{2}{c}{\textbf{MuSiQue}} & \multicolumn{2}{|c}{\textbf{Average}} \\
 & EM & F1 & EM & F1 & EM & F1 & EM & F1 & EM & F1 \\
\midrule
\multicolumn{1}{l}{Naive} & 39.1 & 52.4 & 39.0 & 55.3 & 26.5 & 41.1 & 6.3 & 12.5 & 27.7 & 40.3 \\
\multicolumn{1}{l}{BGE-Reranker} & 39.1 & 52.5 & 42.4 & 59.4 & 30.7 & 45.0 & 7.1 & 13.7 & 29.8 & 42.7 \\
\multicolumn{1}{l}{RankZephyr} & 39.3 & 52.8 & 42.8 & 59.7 & 31.1 & 45.3 & 6.5 & 12.8 & 29.9 & 42.7 \\
\multicolumn{1}{l}{Rank-K} & 39.1 & 52.8 & 42.9 & 60.1 & 31.0 & 44.9 & 7.0 & 13.5 & 30.0 & 42.8 \\
\multicolumn{1}{l}{Provence} & 43.2 & 56.4 & 35.3 & 49.4 & 25.0 & 35.8 & 5.1 & 10.0 & 27.2 & 37.9 \\
\multicolumn{1}{l}{ComPACT} & 19.9 & 26.7 & 41.6 & 58.0 & 32.7 & 45.1 & 4.8 & 10.2 & 24.8 & 35.0 \\
\multicolumn{1}{l}{Refiner} & 27.6 & 36.8 & 40.5 & 56.4 & 30.3 & 43.3 & 6.4 & 12.2 & 26.2 & 37.2 \\
\multicolumn{1}{l}{EXIT} & 40.4 & 51.5 & 29.1 & 41.6 & 17.8 & 27.2 & 3.9 & 9.1 & 22.8 & 32.4 \\ \midrule
\multicolumn{1}{l}{\ours} & \textbf{47.5} & \textbf{61.5} & \textbf{45.9} & \textbf{64.7} & \textbf{46.1} & \textbf{59.4} & \textbf{15.0} & \textbf{24.0} & \textbf{38.6} & \textbf{52.4} \\
\bottomrule
\end{tabular}
\caption{Results on Llama3.2-3B-Instruct with EM and F1 scores.}
\label{tab:multihop_results_llama3b}
\end{table*}

\begin{table*}[t]
\centering
\begin{tabular}{ccccccccc|cc}
\toprule
\multirow{2}{*}{\textbf{Method}} & \multicolumn{2}{c}{\textbf{AmbigQA}} & \multicolumn{2}{c}{\textbf{HotpotQA}} & \multicolumn{2}{c}{\textbf{2WIKI}} & \multicolumn{2}{c}{\textbf{MusiQue}} & \multicolumn{2}{|c}{\textbf{Average}} \\
 & EM & F1 & EM & F1 & EM & F1 & EM & F1 & EM & F1 \\
\midrule
\multicolumn{1}{l}{Naive}        & 8.9  & 22.6 & 9.9  & 22.7 & 9.8  & 22.6 & 3.0  & 9.0 & 7.9 & 19.2 \\
\multicolumn{1}{l}{BGE-Reranker} & 8.7  & 22.6 & 13.9 & 28.7 & 10.8 & 25.4 & 4.0 & 10.6 & 9.4 & 21.8 \\
\multicolumn{1}{l}{RankZephyr}  & 8.9  & 22.7 & 14.0 & 28.7 & 11.5 & 25.9 & 4.3 & 11.0 & 9.7 & 22.1  \\
\multicolumn{1}{l}{Rank-K} & 8.2 & 22.1 & 14.1 & 29.3 & 11.7 & 25.3 & 4.2 & 10.8 & 9.6 & 21.9 \\
\multicolumn{1}{l}{Provence}     & 24.4 & 38.0 & 15.8 & 30.2 & 17.0 & 30.9 & 6.6 & 12.9 & 16.0 & 28.0 \\
\multicolumn{1}{l}{ComPACT}      & 9.3  & 19.4 & 19.1 & 35.6 & 19.4 & 34.4 & 5.6 & 12.2 & 13.4 & 25.4 \\
\multicolumn{1}{l}{Refiner}      & 12.0 & 22.2 & 17.7 & 34.0 & 17.3 & 31.5 & 5.7 & 12.6 & 13.2 & 25.1 \\
\multicolumn{1}{l}{EXIT}         & 13.4 & 28.0 & 11.2 & 25.1 & 10.0 & 23.5 & 4.1 & 10.2 & 9.7 & 21.7 \\ \midrule
\multicolumn{1}{l}{\ours}     &    \textbf{24.9} & \textbf{41.8} & \textbf{22.9} & \textbf{41.1} & \textbf{22.7} & \textbf{40.2} & \textbf{9.8} & \textbf{19.1} & \textbf{20.1} & \textbf{35.6} \\
\bottomrule
\end{tabular}
\caption{Results on Qwen2.5-0.5B-Instruct with EM and F1 scores.}
\end{table*}
\begin{table*}[t]
\centering
\begin{tabular}{ccccccccc|cc}
\toprule
\multirow{2}{*}{\textbf{Method}} & \multicolumn{2}{c}{\textbf{AmbigQA}} & \multicolumn{2}{c}{\textbf{HotpotQA}} & \multicolumn{2}{c}{\textbf{2WIKI}}  & \multicolumn{2}{c}{\textbf{MuSiQue}}  & \multicolumn{2}{|c}{\textbf{Average}} \\
 & EM & F1 & EM & F1 & EM & F1 & EM & F1 & EM & F1 \\
\midrule
\multicolumn{1}{l}{Naive} & 40.7 & 50.6 & 37.2 & 50.8 & 30.9 & 38.3 & 10.1 & 18.0 & 29.7 & 39.4 \\
\multicolumn{1}{l}{BGE-Reranker} & 39.9 & 50.2 & 42.7 & 57.4 & 35.2 & 43.5 & 10.7 & 18.9 & 32.1 & 42.5 \\
\multicolumn{1}{l}{RankZephyr} & 40.9 & 50.9 & 42.0 & 56.5 & 35.8 & 43.7 & 11.1 & 19.0 & 32.5 & 42.5 \\
\multicolumn{1}{l}{Rank-K} & 40.6 & 50.8 & 43.1 & 57.3 & 34.9 & 42.4 & 10.7 & 18.8 & 32.3 & 42.3 \\
\multicolumn{1}{l}{Provence} & 46.4 & 58.3 & 38.6 & 52.8 & 34.9 & 42.4 & 12.2 & 20.1 & 33.0 & 43.4 \\
\multicolumn{1}{l}{ComPACT} & 20.4 & 29.3 & 43.8 & 59.2 & 39.7 & 47.8 & 9.7  & 16.8 & 28.4 & 38.3 \\
\multicolumn{1}{l}{Refiner} & 25.2 & 35.0 & 43.1 & 58.4 & 38.2 & 46.0 & 12.2 & 20.3 & 29.7 & 39.9 \\
\multicolumn{1}{l}{EXIT} & 41.7 & 52.0 & 33.8 & 46.5 & 30.8 & 37.5 & 10.0 & 17.6 & 29.1 & 38.4 \\ \midrule
\multicolumn{1}{l}{\ours} & \textbf{48.8} & \textbf{61.6} & \textbf{46.5} & \textbf{64.0} & \textbf{47.0} & \textbf{57.1} & \textbf{15.8} & \textbf{25.7} & \textbf{39.5} & \textbf{52.1} \\
\bottomrule
\end{tabular}
\caption{Results on Qwen2.5-1.5B-Instruct with EM and F1 scores.}
\label{tab:multihop_results_qwen15b}
\end{table*}
\begin{table*}[t]
\centering
\begin{tabular}{ccccccccc|cc}
\toprule
\multirow{2}{*}{\textbf{Method}} & \multicolumn{2}{c}{\textbf{AmbigQA}} & \multicolumn{2}{c}{\textbf{HotpotQA}} & \multicolumn{2}{c}{\textbf{2WIKI}} & \multicolumn{2}{c}{\textbf{MuSiQue}} & \multicolumn{2}{|c}{\textbf{Average}} \\
 & EM & F1 & EM & F1 & EM & F1 & EM & F1 & EM & F1 \\
\midrule
\multicolumn{1}{l}{Naive}      & 43.8 & 54.5 & 37.2 & 51.4 & 34.8 & 42.3 & 9.4 & 16.3 & 31.3 & 41.1 \\
\multicolumn{1}{l}{BGE-Reranker} & 44.9 & 55.4 & 44.0 & 58.9 & 41.6 & 49.7 & 11.2 & 18.4 & 35.4 & 45.6 \\
\multicolumn{1}{l}{RankZephyr}  & 44.1 & 54.8 & 43.1 & 58.0 & 40.3 & 48.3 & 10.7 & 17.7 & 34.6 & 44.7 \\
\multicolumn{1}{l}{Rank-K} & 44.2 & 55.1 & 42.0 & 57.2 & 38.6 & 46.3 & 10.3 & 17.4 & 33.8 & 44.0 \\
\multicolumn{1}{l}{Provence}     & 47.5 & 58.6 & 40.2 & 54.3 & 37.6 & 44.6 & 12.6 & 20.0 & 34.5 & 44.4 \\
\multicolumn{1}{l}{ComPACT}      & 22.0 & 29.4 & 45.8 & 60.8 & 43.6 & 51.5 & 10.5 & 17.3 & 30.5 & 39.8 \\
\multicolumn{1}{l}{Refiner}      & 29.1 & 37.7 & 44.7 & 59.7 & 41.8 & 49.7 & 12.9 & 20.2 & 32.1 & 41.8 \\
\multicolumn{1}{l}{EXIT}         & 42.4 & 53.6 & 35.6 & 48.3 & 33.1 & 40.0 & 9.4 & 15.9 & 30.1 &  39.5 \\ \midrule
\multicolumn{1}{l}{\ours}       & \textbf{49.9} & \textbf{62.5} & \textbf{49.7} & \textbf{66.8} & \textbf{54.8} & \textbf{64.3} & \textbf{17.8} & \textbf{27.5} & \textbf{43.1} & \textbf{55.3} \\ 
\bottomrule
\end{tabular}
\caption{Results on Qwen2.5-3B-Instruct with EM and F1 scores.}
\label{tab:multihop_results_qwen3b}
\end{table*}
\begin{table*}[t]
\centering
\begin{tabular}{ccccccccc|cc}
\toprule
\multirow{2}{*}{\textbf{Method}} & \multicolumn{2}{c}{\textbf{AmbigQA}} & \multicolumn{2}{c}{\textbf{HotpotQA}} & \multicolumn{2}{c}{\textbf{2WIKI}} & \multicolumn{2}{c}{\textbf{MuSiQue}} & \multicolumn{2}{|c}{\textbf{Average}} \\
 & EM & F1 & EM & F1 & EM & F1 & EM & F1  & EM & F1 \\
\midrule
\multicolumn{1}{l}{Naive}        & 49.1 & 60.8 & 46.6 & 62.1 & 39.4 & 48.7 & 12.4 & 21.2 & 36.9 & 48.2 \\
\multicolumn{1}{l}{BGE-Reranker} & 49.6 & 61.5 & 49.8 & 66.1 & 45.7 & 54.7 & 13.8 & 22.6 & 39.7 & 51.2 \\
\multicolumn{1}{l}{RankZephyr}  & 49.5 & 61.2 & 50.6 & 66.7 & 44.7 & 54.0 & 14.3 & 22.8 & 39.8 & 51.2 \\
\multicolumn{1}{l}{Rank-K} & 49.4 & 60.9 & 51.0 & 67.5 & 49.2 & 59.1 & 13.8 & 22.5 & 40.9 & 52.5 \\
\multicolumn{1}{l}{Provence}     & 48.9 & 61.6 & 43.3 & 58.0 & 39.0 & 47.1 & 13.0 & 21.3 & 36.1 & 47.0 \\
\multicolumn{1}{l}{ComPACT}      & 24.9 & 33.5 & 49.0 & 64.6 & 45.1 & 53.4 & 10.4 & 18.0 & 32.4 & 42.4 \\
\multicolumn{1}{l}{Refiner}      & 33.3 & 43.9 & 47.6 & 63.2 & 45.2 & 53.1 & 13.3 & 21.8 & 34.9 & 45.5 \\
\multicolumn{1}{l}{EXIT}         & 46.7 & 57.8 & 40.1 & 53.8 & 36.5 & 44.4 & 11.5 & 20.0 & 33.7 & 44.0 \\ \midrule
\multicolumn{1}{l}{\ours}         & \textbf{53.8} & \textbf{66.2} & \textbf{51.9} & \textbf{69.6} & \textbf{57.3} & \textbf{67.2} & \textbf{17.4} & \textbf{27.9} & \textbf{45.1} & \textbf{57.7} \\
\bottomrule
\end{tabular}
\caption{Results on Qwen2.5-7B-Instruct with EM and F1 scores.}
\label{tab:multihop_results_qwen7b}
\end{table*}

\end{document}